\pdfoutput=1

\documentclass[11pt]{article}

\usepackage[]{acl}
\usepackage{times}
\usepackage{latexsym}
\usepackage{amsmath}
\usepackage{subcaption}
\usepackage{array}
\usepackage{graphicx}
\usepackage[T1]{fontenc}
\usepackage{hyperref}
\usepackage[utf8]{inputenc}
\usepackage[colorinlistoftodos]{todonotes}
\usepackage{microtype}

\usepackage{textcomp}  
\usepackage{scalerel}  

\usepackage{lipsum}

\title{NoisyAnnot@ Causal News Corpus 2022: Causality Detection using Multiple Annotation Decisions}

\author{Quynh Anh Nguyen\textsuperscript{1, 2} 
Arka Mitra\textsuperscript{2}\\         
\textsuperscript{1} University of Milan $\quad$\textsuperscript{2} ETH Zürich\\
\texttt{quynguyen@ethz.ch, amitra@ethz.ch}
}

\begin{document}
\maketitle
\begin{abstract}
The paper describes the work that has been submitted to the $5^{th}$ workshop on Challenges and Applications of Automated Extraction of socio-political events from text (CASE 2022). The work is associated with Subtask 1 of Shared Task 3 that aims to detect causality in protest news corpus. The authors used different large language models with customized cross-entropy loss functions that exploit annotation information. The experiments showed that \textit{bert-based-uncased} with refined cross-entropy outperformed the others, achieving a F1 score of 0.8501 on the Causal News Corpus dataset.
\end{abstract}

\section{Introduction}
\label{sec:intro}

A causal relationship in a sentence implies an underlying semantic dependency between the two main clauses. The clauses in these sentences are generally connected by markers which can have different parts of tags in the sentence. Moreover, the markers can be either implicit or explicit and for these reasons, one cannot rely on regex or dictionary-based systems. Thus, there is a need to investigate the context of the sentences. For the given task, we exploited different large language models that provide a contextual representation of sentences to tackle causality detection.

Shared task 3 in CASE-2022 \cite{tan-etal-2022-event} aims for causality detection in news corpus, which can be structured as a text classification problem with binary labels. Pre-trained transformer-based models \cite{transformers} have shown success on tackling a wide range of NLP tasks including text generation, text classification, etc. The authors look into inter-annotation agreements and number of experts and how they can be included in the loss to improve the performance of the pre-trained models. 


The main contributions of the paper are as follows:
\begin{enumerate}
    \item Extensive experimentation with different large language models.
    \item Incorporation of additional annotation information, i.e inter-annotation agreement and the number of annotators, to the loss.
\end{enumerate}

The remaining paper is formulated as follows: Section ~\ref{sec:related_work} reviews the related work,  section ~\ref{sec:dataset} describes the dataset on which the work has been done, section~\ref{sec:method} discusses the methodology used in the paper, the following section discusses the results and provides an ablation of the various loss functions introduced and finally, section ~\ref{sec:conclusion} concludes the paper and suggests future works.

\section{Related Work}
\label{sec:related_work}

Multiple annotations on a single sample reduce the chances of the labelling to be incorrect or bias being incorporated into the dataset \cite{Snow2008CheapAF}. Including multiple annotators also leads to disagreement among the labels that have been provided by them. The final or gold annotation is then usually determined by majority voting \cite{Sabou2014CorpusAT} or by using the label of an "expert" \cite{Waseem2016HatefulSO}. There are also different methodologies which do not use majority voting to select the "ground truth".

Expectation Maximization algorithm has been used to account for the annotator error \cite{Dawid1979MaximumLE}. Entropy metrics have been developed to identify the performance of the annotators\cite{Waterhouse2012PayBT, Hovy2013LearningWT, Gordon2021TheDD}. Multi-task learning is also used to deal with disagreement in the labels \cite{Fornaciari2021BeyondB, Liu2019MultiTaskDN, Cohn2013ModellingAB, Davani2022DealingWD}. There are methods which include the annotation disagreement into the loss function for part of speech tagging \cite{Plank2014LearningPT, Prabhakaran2012StatisticalMT} on SVMs and perceptron model. The present work considers the inter-annotator agreement as well as the number of annotators into the loss function for any model. The work also compares the performance when the annotators who disagree with the majority voting has been ignored.

\section{Dataset}
\label{sec:dataset}
\begin{figure}[h!]
    \centering
    \includegraphics[scale=0.58]{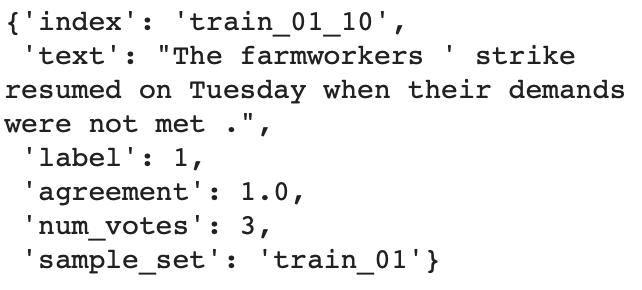}
    \caption{A datapoint from the provided training data.}
    \label{fig:datapoint}
\end{figure}

    The Causal News Corpus dataset \cite{tan_paper_dataset} consists of 3,559 event sentences extracted from protest event news. Each sample in the dataset contains the text, the corresponding label, the number of experts who annotated the label and the degree of agreement among the experts. Figure~\ref{fig:datapoint} shows a sample from the provided training set. The training data is fairly balanced, containing 1603 sentences with a causal structure and 1322 sentences without a causal structure. Also, the number of causal and non-causal sentences in the validation set does not differ significantly. Finally, 311 news articles have been used as test set for evaluation.  

Besides the binary labels, the Causal News Corpus dataset also provides additional information regarding the number of experts who labeled the sentence and the percentage of agreement between them. Figure \ref{fig:datapoint} shows that the number of experts who annotated the text \textit{"The farmworkers'strike resumed on Tuesday when their demands were not met."} is 3 ($\texttt{num\_votes}=3$). Also, all of the experts labeled the sentence to be causal so the agreement is 1.0 (100\% agreement) and the label is 1. In case only one of three experts assigned label 1 to the previous text, the three predictors \texttt{num\_votes}, \texttt{agreement}, \texttt{label} would now become $3$, $\frac{2}{3}$, $0$ respectively.
In this paper, the authors exploit this information to give the model more prior and thus potentially improve the model's performance, which has been described in more detail in section \ref{sec:method}.
\section{Methodology}
\label{sec:method}
The section discusses the pipeline, the different types of loss functions that were implemented, and the experimental details that have been used in the third shared task for CASE 2022 \cite{tan-etal-2022-event}.

\subsection{Pipeline}
    The authors finetuned large language models with different loss functions to tackle Subtask 1 in Shared Task 3 of CASE@EMNLP-2022, causality detection in a given sentence. The problem can be reformulated as a binary classification where the model predicts whether the sentence is causal or not.
Since contextual awareness plays an essential role in handling this specific task, the authors used several transformer-based models, namely,
 BERT \cite{Devlin2019BERTPO}, FinBERT \cite{Liu2020FinBERTAP}, XLNET \cite{Yang2019XLNetGA} and RoBERTa \cite{Liu2019RoBERTaAR}.

 The given sentence is first tokenized by a tokenizer from the corresponding pretrained model architecture provided by HuggingFace \cite{Wolf2019HuggingFacesTS}. The vector output from the tokenization stage is then fed as input to the model. The most informative token is the classification token ([CLS]), which is a special token that can be used as a sentence representation. The [CLS] token is then passed through a feed-forward network to generate logits. The softmax over the logits gives us the probability of whether the sentence is causal or not. For each model, the authors experimented with cross-entropy loss and proposed two loss functions described in detail in subsection \ref{subsec:Loss}.
 
\subsection{Loss Functions} \label{subsec:Loss}
    \paragraph{Cross Entropy Loss} The loss of the classification task can be represented by a simple cross-entropy loss, as shown in Equation \ref{eq:basic_ce}:
    \begin{equation}
    \label{eq:basic_ce}
    \begin{aligned}
        L &=  \frac{1}{M} \sum_{i=1}^M (-y^{true}_ilog(y^{pred}_i)\\
            & - (1-y^{true}_i)log(1-y^{pred}_i))
    \end{aligned}
    \end{equation}
    where $y_i^{true}$ and $y_i^{pred}$ denote the true label and the predicted label for the $i^{th}$ input in a batch of M sentences.
    \paragraph{Noisy Cross Entropy Loss} The dataset not only provides the standard information about \texttt{\{text, label\}}, but also contains the information about the number of experts who annotated the sentence's label, and proportion of agreement between them. The authors have considered the annotation by each of the experts to be the true label for the sentence. For a sentence with $n$ expert annotations ($\texttt{num\_votes}=n$) and $r$ percent of agreement ($\texttt{agreement}=r$), the loss for each sentence can be written as shown in Equation \ref{eq:single_sent}.
\begin{equation} \label{eq:single_sent}
L=\begin{cases}
(-r log(y^{pred})& \\
-(1-r) log(1-y^{pred})), & \text{if }y^{true}= 1\,,  \\
(-(1-r) log(y^{pred})& \\
-r log(1-y^{pred})), & \text{if }y^{true}= 0\,.
\end{cases}
\end{equation}
The equations can be combined and the loss for a batch of M sentences can be rewritten as:
\begin{equation} \label{eq:batch_eqn}
    \begin{aligned}
         L = &\frac{1}{\sum_{i=1}^M n_i} \sum_{i=1}^M (-y^{true}_i n_i(r_ilog(y^{pred}_i)&\\
         & +(1-r_i)log(1-y^{pred}_i))&\\
     & - (1-y^{true}_i) n_i(r_ilog(1-y^{pred}_i)&\\
     &+(1-r_i)log(y^{pred}_i)))
    \end{aligned}
\end{equation}.

\begin{figure}[h]
    \centering
    \includegraphics[scale=0.55]{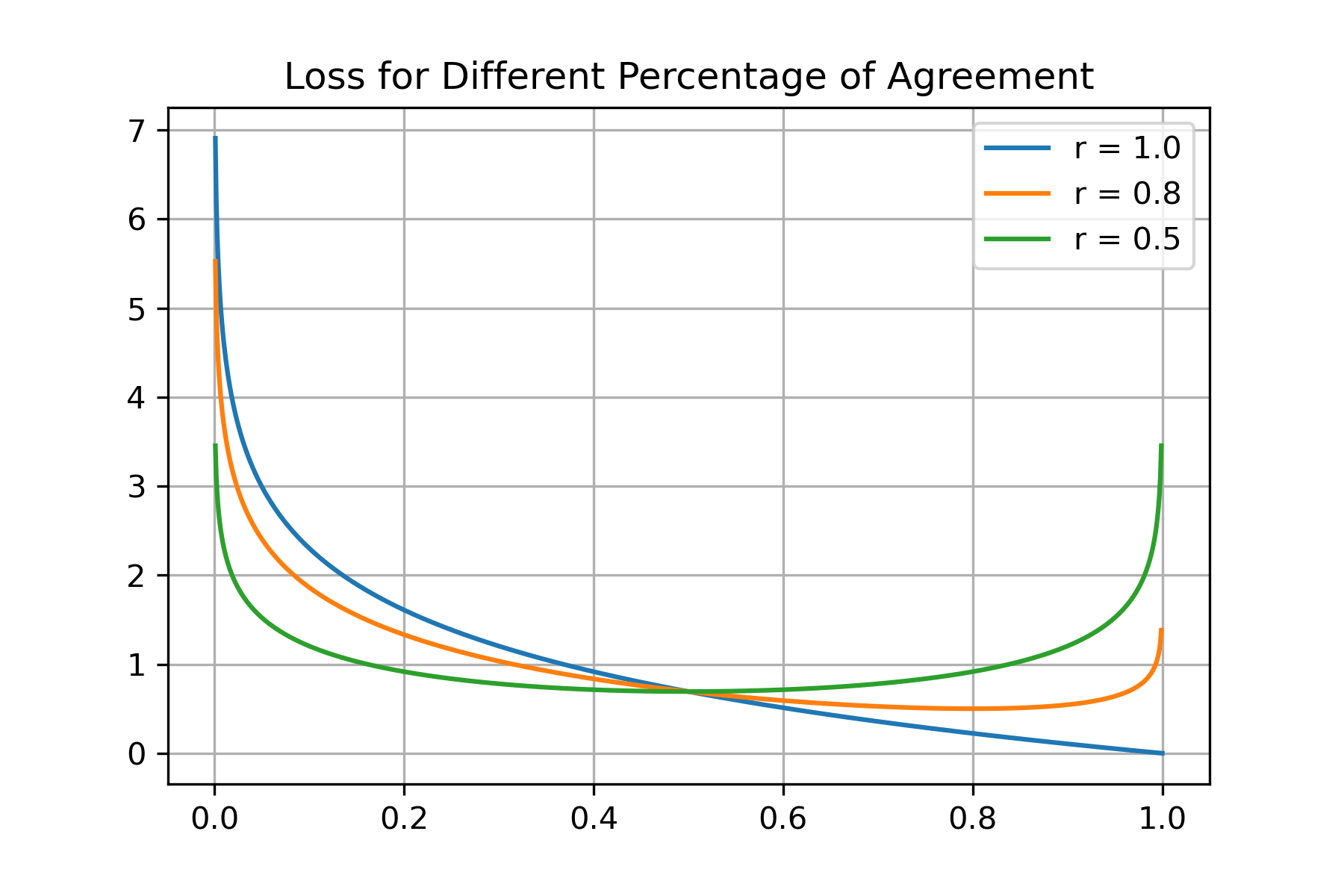}
    \caption{Loss for noisy cross-entropy}
    \label{fig:eqn_max}
\end{figure}

The different annotations from all the experts has been considered, adding more information to the model. Equation ~\ref{eq:batch_eqn} takes the $n$ votes from the different experts into account, out of which $n\times r$ times it is assigned the correct label, and the incorrect label has been used the other $n\times (1-r)$ times. If the labels from the different experts are taken directly, there will be conflicts in the labels when the experts disagree. Considering the loss for one sentence when the true label is 1, the derivative of the loss is shown in Equation \ref{eq:deriv}. Figure \ref{fig:eqn_max} shows that the loss is minimized when $y^{pred}$ is equal to $r$ and its minima shifts from 1 to 0 as the level of agreement decreases when the true label is 1. A similar profile is obtained when the true label is considered to be 0. The formulation pushes the solution to a distribution where the ideal output is not a one-hot encoding, which is similar to the label smoothing method. Label smoothing was initially proposed by Szegedy et al. \cite{smooth-labeling} to improve the performance of the Inception architecture on the ImageNet dataset \cite{Deng2009ImageNetAL}. In label-smoothing, the ground truth sent to the model is not encoded as a one-hot representation. Since there are conflicts in the annotations and the loss considers all of the noisy data, it has been referred as noisy cross-entropy loss.
\begin{equation}
    \frac{\partial L}{\partial y^{pred}} = \frac{y^{pred}-r}{y^{pred}(1-y^{pred})}
    \label{eq:deriv}
\end{equation}

\paragraph{Refined Cross Entropy Loss} The ideal output of the model should be close to the ground truth label. Thus, a modification to loss function should be done to improve the performance. The error occurs when the annotators who have not agreed for a particular label have also been taken into consideration. The number of experts who provided the correct label can also be an important signal to the model. If a sentence has been given a label by a more significant number of experts, the model should be penalized more if the sentence is misclassified. The new loss, over a batch of M sentences, can thus be written as :
\begin{equation} \label{eq:batch_eqn_final}
    \begin{aligned}
         L =  &\frac{1}{\sum_{i=1}^M n_ir_i} \sum_{i=1}^M (-y^{true}_i n_ir_ilog(y^{pred}_i)&\\
     & - (1-y^{true}_i) n_ir_ilog(1-y^{pred}_i))
    \end{aligned}
\end{equation}
.

The number of causal and non-causal sentences is almost the same and there is no significant class imbalance. The authors have thus not considered weight penalization to the class with the higher number of samples.

\subsection {Experimental Details}
     The experiments have been performed in PyTorch \cite{pytorch} and the authors used the HuggingFace \cite{Wolf2019HuggingFacesTS} library to generate the pipeline for the different experiments. Each model has been trained for 10 epochs with a learning rate of $5\times 10^{-5}$ and a seed of 42 for reproducibility. Various models have been considered and trained with the same set of hyperparameters. The code is made publicly available on Github \footnote{\url{https://github.com/jyanqa/case-2022-causual-event}}.

\begin{table*}[tph]
    \begin{center}
        \begin{tabular}{ | m{7cm} | m{2cm}| m{2cm} | m{2cm}|}
        \hline
          \textbf{Model name} & \textbf{Cross \quad \quad Entropy} & \textbf{Noisy \quad \quad Cross \quad \quad Entropy} &\textbf{Refined Cross \quad \quad Entropy}\\
          \hline
           bert-based-cased \cite{Devlin2019BERTPO} & \textbf{0.8251} & 0.8225 & 0.8235\\
           bert-base-uncased \cite{Devlin2019BERTPO}& 0.8283 & 0.8313 & \textbf{0.8501}\\
           bert-large-cased \cite{Devlin2019BERTPO}& 0.7105 & \textbf{0.7549}  & 0.7105\\
         xlnet-based-cased \cite{Yang2019XLNetGA}& 0.7953& \textbf{0.8216} & 0.8199\\
         roberta-base \cite{Liu2019RoBERTaAR}& 0.8279 & 0.8279& \textbf{0.8280}\\
         \hline
        \end{tabular}
         \caption{Evaluation of models on different loss functions. The best F1 score of each model is marked in bold. }
        \label{tab:celoss_results}
    \end{center}
\end{table*}
\section{Results and Discussion}
\label{sec:results}

In this section, the results of the different models and the different losses are discussed. 

Table~\ref{tab:celoss_results} shows the evaluation of the different models on the validation set. Performances of four in five models, excepting the \textit{bert-base-uncased} case, are enhanced by leveraging the modified cross-entropy loss. In fact, the F1 scores of four models are significantly increasing when we replaced vanilla cross-entropy loss with noisy cross-entropy loss and refined cross-entropy loss. Specifically, model fine-tuned from \textit{bert-base-uncased} investigating Refined cross-entropy loss function yields the best performance in all experimented models with F1 score of 0.8501. On the other hand, \textit{bert-base-cased} is the only pretrained model that does not benefit from customized cross-entropy losses. Adapting vanilla cross-entropy function on \textit{bert-base-cased} model results in its best F1 scores of 0.8251. 

The models with noisy and refined cross-entropy loss utilizes the annotated information and thus performs better. The noisy cross-entropy loss is similar to restricting the highest probability output that a model can predict. However, in almost all cases, the degree of agreement was either 1 or $\frac{2}{3}$. In general, the smooth labelling has a value in the range of 0.9 to 1. Different contradicting annotations of labels might make the model face difficulties in learning and yielding an accurate prediction for each sentence. The refined cross-entropy solely considers the labels that do not contradict each other, thus it performs the best. 

Moreover, the experiments show that \textit{roberta-based} models achieve lower performance compared to BERT-based models, especially \textit{bert-base-uncased} models. The model pretrained on \textit{bert-large-cased} has been fine-tuned for only one epoch due to computation limitations. Their F1 scores are worse than those of \textit{bert-base-cased} and \textit{bert-base-uncased} models. \textit{bert-base} models result in better performance, as compared to models fine-tuned on  \textit{roberta-base}. The reason could be that RoBERTa-based models had not been trained on next sentence prediction (NSP) while BERT-based models were. Causality detection can benefit from NSP. A sentence can be considered to be two relevant clauses that are joined by a causal effect. Thus, knowing if the clauses are relevant or not benefits the task of causality detection.
 
\begin{figure}[hp]
\centering
\begin{subfigure}{0.23\textwidth}
    \includegraphics[width=\textwidth]{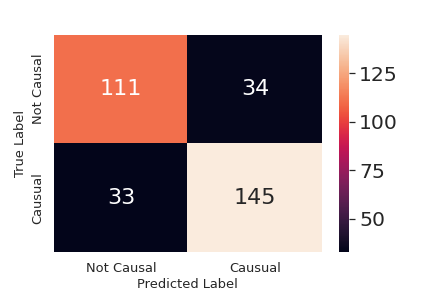}
    \caption{Vanilla CE}
    \label{fig:first}
\end{subfigure}
\hfill
\begin{subfigure}{0.23\textwidth}
    \includegraphics[width=\textwidth]{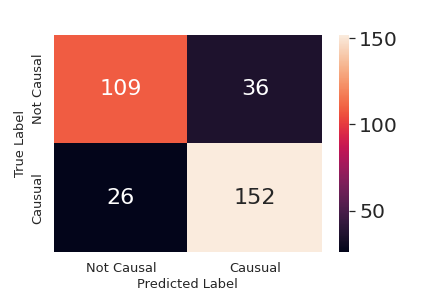}
    \caption{Noisy CE}
    \label{fig:second}
\end{subfigure}
\hfill
\begin{subfigure}{0.23\textwidth}
    \includegraphics[width=\textwidth]{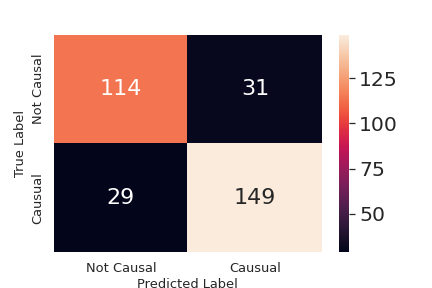}
    \caption{Refined CE}
    \label{fig:third}
\end{subfigure}
        
\caption{Confusion matrix for the different losses}
\label{fig:error_analysis}
\end{figure}


Figure \ref{fig:error_analysis} shows the confusion matrix resulting from \textit{bert-base-uncased} models which result the best F1 scores in all implemented models. Models are generally good at predicting non-causal sentences regardless of the loss function used. In fact, true negatives and true positives are always the highest measures compared to the others. On the other hand, there is a clear trend in the number of true positives when we shift the loss function from vanilla to noisy and refined cross-entropy. In particular, the model yields 145 true positives and is improved to 152 and 149 true positives when we replaced vanilla cross-entropy loss with noisy and refined cross-entropy loss function.
            
\section{Conclusion}
\label{sec:conclusion}
This paper presents our work on detecting causal effect relationships in news corpus by fine-tuning Transformers-based models and adapting multiple loss functions. The experiments showed that considering annotation information using customized loss functions significantly improved the model performance in four out of five experimented models. Besides, the experiments show that BERT outperformed RoBERTa, which can be attributed to the fact that RoBERTa is not trained on NSP. Last but not least, the \textit{bert-base-uncased} obtained the best performance amongst all 15 models with an F1-score of 0.8501 in validation set and 84.930 in the test set using the refined cross-entropy loss that takes account of the annotation information presented in the dataset.

The authors plan to look into exploiting the uncertainty of the annotator's information and parameterizing the loss function to further enhance the model's performance. 
\bibliography{anthology,custom}

\begin{thebibliography}{24}
\expandafter\ifx\csname natexlab\endcsname\relax\def\natexlab#1{#1}\fi

\bibitem[{Cohn and Specia(2013)}]{Cohn2013ModellingAB}
Trevor Cohn and Lucia Specia. 2013.
\newblock Modelling annotator bias with multi-task gaussian processes: An
  application to machine translation quality estimation.
\newblock In \emph{ACL}.

\bibitem[{Davani et~al.(2022)Davani, D'iaz, and
  Prabhakaran}]{Davani2022DealingWD}
Aida~Mostafazadeh Davani, Mark D'iaz, and Vinodkumar Prabhakaran. 2022.
\newblock Dealing with disagreements: Looking beyond the majority vote in
  subjective annotations.
\newblock \emph{Transactions of the Association for Computational Linguistics},
  10:92--110.

\bibitem[{Dawid and Skene(1979)}]{Dawid1979MaximumLE}
A.~Philip Dawid and Allan Skene. 1979.
\newblock Maximum likelihood estimation of observer error‐rates using the em
  algorithm.
\newblock \emph{Journal of The Royal Statistical Society Series C-applied
  Statistics}, 28:20--28.

\bibitem[{Deng et~al.(2009)Deng, Dong, Socher, Li, Li, and
  Fei-Fei}]{Deng2009ImageNetAL}
Jia Deng, Wei Dong, Richard Socher, Li-Jia Li, K.~Li, and Li~Fei-Fei. 2009.
\newblock Imagenet: A large-scale hierarchical image database.
\newblock \emph{2009 IEEE Conference on Computer Vision and Pattern
  Recognition}, pages 248--255.

\bibitem[{Devlin et~al.(2019)Devlin, Chang, Lee, and
  Toutanova}]{Devlin2019BERTPO}
Jacob Devlin, Ming-Wei Chang, Kenton Lee, and Kristina Toutanova. 2019.
\newblock \href {https://doi.org/10.18653/v1/N19-1423} {{BERT}: Pre-training of
  deep bidirectional transformers for language understanding}.
\newblock In \emph{Proceedings of the 2019 Conference of the North {A}merican
  Chapter of the Association for Computational Linguistics: Human Language
  Technologies, Volume 1 (Long and Short Papers)}, pages 4171--4186,
  Minneapolis, Minnesota. Association for Computational Linguistics.

\bibitem[{Fornaciari et~al.(2021)Fornaciari, Uma, Paun, Plank, Hovy, and
  Poesio}]{Fornaciari2021BeyondB}
Tommaso Fornaciari, Alexandra Uma, Silviu Paun, Barbara Plank, Dirk Hovy, and
  Massimo Poesio. 2021.
\newblock Beyond black \& white: Leveraging annotator disagreement via
  soft-label multi-task learning.
\newblock In \emph{NAACL}.

\bibitem[{Gordon et~al.(2021)Gordon, Zhou, Patel, Hashimoto, and
  Bernstein}]{Gordon2021TheDD}
Mitchell~L. Gordon, Kaitlyn Zhou, Kayur Patel, Tatsunori~B. Hashimoto, and
  Michael~S. Bernstein. 2021.
\newblock The disagreement deconvolution: Bringing machine learning performance
  metrics in line with reality.
\newblock \emph{Proceedings of the 2021 CHI Conference on Human Factors in
  Computing Systems}.

\bibitem[{Hovy et~al.(2013)Hovy, Berg-Kirkpatrick, Vaswani, and
  Hovy}]{Hovy2013LearningWT}
Dirk Hovy, Taylor Berg-Kirkpatrick, Ashish Vaswani, and Eduard~H. Hovy. 2013.
\newblock Learning whom to trust with mace.
\newblock In \emph{NAACL}.

\bibitem[{Liu et~al.(2019)Liu, He, Chen, and Gao}]{Liu2019MultiTaskDN}
Xiaodong Liu, Pengcheng He, Weizhu Chen, and Jianfeng Gao. 2019.
\newblock Multi-task deep neural networks for natural language understanding.
\newblock In \emph{ACL}.

\bibitem[{Liu et~al.(2020)Liu, Huang, Huang, Li, and Zhao}]{Liu2020FinBERTAP}
Zhuang Liu, Degen Huang, Kaiyu Huang, Zhuang Li, and Jun Zhao. 2020.
\newblock Finbert: A pre-trained financial language representation model for
  financial text mining.
\newblock In \emph{IJCAI}.

\bibitem[{Paszke et~al.(2019)Paszke, Gross, Massa, Lerer, Bradbury, Chanan,
  Killeen, Lin, Gimelshein, Antiga, Desmaison, Kopf, Yang, DeVito, Raison,
  Tejani, Chilamkurthy, Steiner, Fang, Bai, and Chintala}]{pytorch}
Adam Paszke, Sam Gross, Francisco Massa, Adam Lerer, James Bradbury, Gregory
  Chanan, Trevor Killeen, Zeming Lin, Natalia Gimelshein, Luca Antiga, Alban
  Desmaison, Andreas Kopf, Edward Yang, Zachary DeVito, Martin Raison, Alykhan
  Tejani, Sasank Chilamkurthy, Benoit Steiner, Lu~Fang, Junjie Bai, and Soumith
  Chintala. 2019.
\newblock \href
  {http://papers.neurips.cc/paper/9015-pytorch-an-imperative-style-high-performance-deep-learning-library.pdf}
  {Pytorch: An imperative style, high-performance deep learning library}.
\newblock In \emph{Advances in Neural Information Processing Systems 32}, pages
  8024--8035. Curran Associates, Inc.

\bibitem[{Plank et~al.(2014)Plank, Hovy, and S{\o}gaard}]{Plank2014LearningPT}
Barbara Plank, Dirk Hovy, and Anders S{\o}gaard. 2014.
\newblock Learning part-of-speech taggers with inter-annotator agreement loss.
\newblock In \emph{EACL}.

\bibitem[{Prabhakaran et~al.(2012)Prabhakaran, Bloodgood, Diab, Dorr, Levin,
  Piatko, Rambow, and Durme}]{Prabhakaran2012StatisticalMT}
Vinodkumar Prabhakaran, Michael Bloodgood, Mona~T. Diab, B.~Dorr, Lori~S.
  Levin, Christine~D. Piatko, Owen Rambow, and Benjamin~Van Durme. 2012.
\newblock Statistical modality tagging from rule-based annotations and
  crowdsourcing.
\newblock In \emph{ExProM@ACL}.

\bibitem[{Sabou et~al.(2014)Sabou, Bontcheva, Derczynski, and
  Scharl}]{Sabou2014CorpusAT}
Marta Sabou, Kalina Bontcheva, Leon Derczynski, and Arno Scharl. 2014.
\newblock Corpus annotation through crowdsourcing: Towards best practice
  guidelines.
\newblock In \emph{LREC}.

\bibitem[{Snow et~al.(2008)Snow, O'Connor, Jurafsky, and Ng}]{Snow2008CheapAF}
Rion Snow, Brendan~T. O'Connor, Dan Jurafsky, and A.~Ng. 2008.
\newblock Cheap and fast – but is it good? evaluating non-expert annotations
  for natural language tasks.
\newblock In \emph{EMNLP}.

\bibitem[{Szegedy et~al.(2016)Szegedy, Vanhoucke, Ioffe, Shlens, and
  Wojna}]{smooth-labeling}
Christian Szegedy, Vincent Vanhoucke, Sergey Ioffe, Jon Shlens, and Zbigniew
  Wojna. 2016.
\newblock \href {https://doi.org/10.1109/CVPR.2016.308} {Rethinking the
  inception architecture for computer vision}.
\newblock In \emph{2016 IEEE Conference on Computer Vision and Pattern
  Recognition (CVPR)}, pages 2818--2826.

\bibitem[{Tan et~al.(2022{\natexlab{a}})Tan, Hürriyetoğlu, Caselli, Oostdijk,
  Hettiarachchi, Nomoto, Uca, and Liza}]{tan-etal-2022-event}
Fiona~Anting Tan, Ali Hürriyetoğlu, Tommaso Caselli, Nelleke Oostdijk, Hansi
  Hettiarachchi, Tadashi Nomoto, Onur Uca, and Farhana~Ferdousi Liza.
  2022{\natexlab{a}}.
\newblock Event causality identification with causal news corpus - shared task
  3, {CASE} 2022.
\newblock In \emph{Proceedings of the 5th Workshop on Challenges and
  Applications of Automated Extraction of Socio-political Events from Text
  (CASE 2022)}, Online. Association for Computational Linguistics.

\bibitem[{Tan et~al.(2022{\natexlab{b}})Tan, Hürriyetoğlu, Caselli, Oostdijk,
  Nomoto, Hettiarachchi, Ameer, Uca, Liza, and Hu}]{tan_paper_dataset}
Fiona~Anting Tan, Ali Hürriyetoğlu, Tommaso Caselli, Nelleke Oostdijk,
  Tadashi Nomoto, Hansi Hettiarachchi, Iqra Ameer, Onur Uca, Farhana~Ferdousi
  Liza, and Tiancheng Hu. 2022{\natexlab{b}}.
\newblock \href {https://aclanthology.org/2022.lrec-1.246} {The causal news
  corpus: Annotating causal relations in event sentences from news}.
\newblock In \emph{Proceedings of the Language Resources and Evaluation
  Conference}, pages 2298--2310, Marseille, France. European Language Resources
  Association.

\bibitem[{Vaswani et~al.(2017)Vaswani, Shazeer, Parmar, Uszkoreit, Jones,
  Gomez, Kaiser, and Polosukhin}]{transformers}
Ashish Vaswani, Noam Shazeer, Niki Parmar, Jakob Uszkoreit, Llion Jones,
  Aidan~N Gomez, \L~ukasz Kaiser, and Illia Polosukhin. 2017.
\newblock \href
  {https://proceedings.neurips.cc/paper/2017/file/3f5ee243547dee91fbd053c1c4a845aa-Paper.pdf}
  {Attention is all you need}.
\newblock In \emph{Advances in Neural Information Processing Systems},
  volume~30. Curran Associates, Inc.

\bibitem[{Waseem and Hovy(2016)}]{Waseem2016HatefulSO}
Zeerak Waseem and Dirk Hovy. 2016.
\newblock Hateful symbols or hateful people? predictive features for hate
  speech detection on twitter.
\newblock In \emph{NAACL}.

\bibitem[{Waterhouse(2012)}]{Waterhouse2012PayBT}
Tamsyn~P. Waterhouse. 2012.
\newblock Pay by the bit: an information-theoretic metric for collective human
  judgment.
\newblock \emph{Proceedings of the 2013 conference on Computer supported
  cooperative work}.

\bibitem[{Wolf et~al.(2020)Wolf, Debut, Sanh, Chaumond, Delangue, Moi, Cistac,
  Rault, Louf, Funtowicz, Davison, Shleifer, von Platen, Ma, Jernite, Plu, Xu,
  Le~Scao, Gugger, Drame, Lhoest, and Rush}]{Wolf2019HuggingFacesTS}
Thomas Wolf, Lysandre Debut, Victor Sanh, Julien Chaumond, Clement Delangue,
  Anthony Moi, Pierric Cistac, Tim Rault, Remi Louf, Morgan Funtowicz, Joe
  Davison, Sam Shleifer, Patrick von Platen, Clara Ma, Yacine Jernite, Julien
  Plu, Canwen Xu, Teven Le~Scao, Sylvain Gugger, Mariama Drame, Quentin Lhoest,
  and Alexander Rush. 2020.
\newblock \href {https://doi.org/10.18653/v1/2020.emnlp-demos.6} {Transformers:
  State-of-the-art natural language processing}.
\newblock In \emph{Proceedings of the 2020 Conference on Empirical Methods in
  Natural Language Processing: System Demonstrations}, pages 38--45, Online.
  Association for Computational Linguistics.

\bibitem[{Yang et~al.(2019)Yang, Dai, Yang, Carbonell, Salakhutdinov, and
  Le}]{Yang2019XLNetGA}
Zhilin Yang, Zihang Dai, Yiming Yang, Jaime~G. Carbonell, Ruslan Salakhutdinov,
  and Quoc~V. Le. 2019.
\newblock Xlnet: Generalized autoregressive pretraining for language
  understanding.
\newblock In \emph{NeurIPS}.

\bibitem[{Zhuang et~al.(2021)Zhuang, Wayne, Ya, and Jun}]{Liu2019RoBERTaAR}
Liu Zhuang, Lin Wayne, Shi Ya, and Zhao Jun. 2021.
\newblock \href {https://aclanthology.org/2021.ccl-1.108} {A robustly optimized
  {BERT} pre-training approach with post-training}.
\newblock In \emph{Proceedings of the 20th Chinese National Conference on
  Computational Linguistics}, pages 1218--1227, Huhhot, China. Chinese
  Information Processing Society of China.

\end{thebibliography}
\bibliographystyle{acl_natbib}


\end{document}